# Using Large Language Models to Measure Symptom Severity in Patients At Risk for Schizophrenia

Andrew X. Chen[1], Guillermo Horga[1], Sean Escola[1,2]

[1]Department of Psychiatry, Columbia University and New York State Psychiatric Institute
[2]Protocol Labs

**Abstract**

Patients who are at clinical high risk (CHR) for schizophrenia need close monitoring of their symptoms to inform appropriate treatments. The Brief Psychiatric Rating Scale (BPRS) is a validated, commonly used research tool for measuring symptoms in patients with schizophrenia and other psychotic disorders; however, it is not commonly used in clinical practice as it requires a lengthy structured interview. Here, we utilize large language models (LLMs) to predict BPRS scores from clinical interview transcripts in 409 CHR patients from the Accelerating Medicines Partnership Schizophrenia (AMP-SCZ) cohort. Despite the interviews not being specifically structured to measure the BPRS, the zero-shot performance of the LLM predictions compared to the true assessment (median concordance: 0.84, ICC: 0.73) approaches human inter- and intra-rater reliability. We further demonstrate that LLMs have substantial potential to improve and standardize the assessment of CHR patients via their accuracy in assessing the BPRS in foreign languages (median concordance: 0.88, ICC: 0.70), and integrating longitudinal information in a one-shot or few-shot learning approach.

# Introduction

Schizophrenia is a severe psychiatric disorder affecting 1% of the population, and is characterized by hallucinations, delusions, and negative symptoms which typically result in marked loss of functioning[1]. Close monitoring and early intervention in patients at clinical high risk (CHR) for schizophrenia may improve outcomes and reduce disease morbidity. An ongoing multi-institutional study of CHR patients is the Accelerating Medicines Partnership Schizophrenia (AMP-SCZ) cohort, which provides longitudinal multi-modal clinical data as well as a large-scale repository of clinical interview transcripts in this patient population[2].

Various standardized scales and risk calculators have been used in CHR populations, with the goal of tracking disease progression and, ideally, foretelling the transition to psychosis[3,4]. The AMP-SCZ project includes several of these scales, and also developed a novel instrument called the PSYCHS (Positive Symptoms and Diagnostic Criteria for the CAARMS[5] Harmonized with the SIPS[6]), though the PSYCHS has yet to be validated[7]. In this work we chose to analyze the metric in this cohort with the most complete data as well as previous validation/reliability studies, the 24-item Brief Psychiatric Rating Scale (BPRS). The BPRS is a widely used research instrument, validated in patients with schizophrenia and other psychotic conditions, that captures a wide range of symptom domains, including positive, negative, and affective symptoms[8]. Within the CHR population, the BPRS has been used for predicting conversion to psychosis, monitoring symptomatology, and assessing treatment response[9–14]. However, its use is limited in many clinical settings due to the time burden required to administer the structured interview[15].

As a promising emerging technology to address this problem, large language models (LLMs) provide a novel and valuable opportunity to predict clinical diagnosis, status, and outcomes from a range of data modalities[16]. The transformer architecture of modern LLMs enables them to capture contextual cues, integrate a wide expanse of background knowledge, seamlessly work between foreign languages, and infer subtle meanings from text with high accuracy. Newer models, such as OpenAI's o-series, are further able to use reasoning and internal chain-of-thought to approach complex problems[17]. Several other studies have already demonstrated success in using LLMs on clinical interview transcripts to accurately predict symptom severity in other psychiatric conditions, mainly in the Distress Analysis Interview Corpus (DAIC) cohort for depression[18–20], and also for anxiety symptoms[21]. Although some studies have used LLMs to quantify linguistic markers of psychosis[22,23], to our knowledge, this work represents the first use of LLMs for predicting clinical severity scores in psychosis and in the CHR population.

Here we apply LLMs to process redacted clinical interview transcripts from the AMP-SCZ dataset and assess its ability to predict BPRS scores. We show that, despite the fact that the interviews were not explicitly designed to assess the BPRS, and without prior training on this dataset, that the LLM is able to infer symptom severity scores that approach the accuracy of human inter- and intra-rater reliability, and have further capabilities that may improve and standardize our ability to monitor symptoms in the CHR population.

## Methods

*Ethics*
A determination was obtained from the Columbia University IRB that this research is <u>not</u> Human Subject Research due to it being a retrospective analysis of a fully de-identified dataset (Protocol #AAAV3944). All research was conducted in accordance with the Declaration of Helsinki.

*Dataset*
Data from a total of n=901 patients were obtained from the AMP-SCZ project of the NIMH Data Archive (NDA) portal after obtaining the appropriate permissions. Patients were followed longitudinally for a varying number of single-day encounters, and during each encounter a subset of the following three modalities were captured (among others):

1) BPRS assessment (Extended 24-item version)
2) "Open" interview
3) "PSYCHS" interview

We only analyzed encounters that had both a BPRS assessment as well as an open and/or PSYCHS interview. In total, that corresponded to 308 open transcripts, 433 PSYCHS transcripts, and 651 BPRS measurements across 409 patients.

Of these, 348 patients were English-speaking, resulting in 238 open and 374 PSYCHS transcripts which were used for the primary analysis. Of the remaining 129 non-English transcripts, the breakdown of languages were:
    Spanish 57
    Korean 39
    Italian 9
    Danish 10
    Cantonese 6
    French 3

For patients with longitudinal data, when there were encounters that included both an open and PSYCHS interview on the same day, we used the PSYCHS interview for that timepoint.

*Large Language Model*
We used the OpenAI model o3-mini-2025-01-31 for the primary analysis, using its default parameters. We accessed the model via the OpenAI Python API. The clinical transcripts were unedited and directly fed into the model as the user input. Instructions and background were placed in the system instructions section. Structured output was utilized, such that the output needed to be a JSON format consisting of the names of the 24 subscores, an explanation for each subscore, and the numerical symptom rating for that subscore.

*Prompt Engineering*
It quickly became apparent that the LLM could not reliably list the 24 symptoms of the Extended 24-item version of the BPRS given its knowledge base. Therefore, we included the full instruction set for the BPRS[24]. One issue is that these instructions omitted that the numerical rating of 1 for any subscore indicated that the symptom was not present; therefore we added that as an option for each symptom. We then put simple instructions for the LLM to provide subscore estimates and explanations at the top and bottom of the system instructions. The full text of the instructions provided in the system instructions can be found in the Supplementary Materials.

*Data Analysis*
The results were analyzed in Python. The main metrics of accuracy used were the same as the ones from Hafkenscheid *et al.* (1993) in which inter-rater and intra-rater reliability were assessed[25]. Briefly, concordance, as defined in Hafkenscheid *et al.,* refers to the fraction of subscores in which the two ratings differ at most by 1 point. Of note, Hafkenscheid *et al.* did not define the specific type of Intraclass Correlation Coefficient (ICC) that they calculated. In this work, we report the (3, k) form of the ICC as calculated by the pingouin Python package, as the intended use case of the LLM (averages of repeated measures for a given transcript) is most similar to this form[26]. To compute the estimated standard error of the RMSE, a bootstrapping approach was used. A bootstrap distribution was created via sampling with replacement a sample with the same size as the original dataset 1000 times. The estimated standard error is the standard deviation of this bootstrap distribution.

*Longitudinal Analysis*
In patients with multiple paired transcripts and scores across time, we provided the LLM different subsets of these data points and assessed their relative performances. The nomenclature for these different subsets is as follows:
- t refers to number of time points ago, where t = 0 is the current (most recent) encounter, t = –1 is the previous encounter, and t = –2 is the one before that
- $x_t$ is the transcript at time point t, and $x_0$ refers to the most recent transcript
- $s_t$ is the BPRS score at time point t, and $s_0$ refers to the most recent score
- the LLM is abstracted as a function f that takes in a subset of the above data points and produces an estimated current score $\hat{s}_0 = f(…)$
- The key measure of accuracy is then the RMSE = $\sqrt{E[(s_0 - \hat{s}_0)^2]}$

In our 153 patients with 2+ time points, one-shot learning corresponds to the case
- Model 1-shot: $\hat{s}_0 = f(x_0, [x_{-1}, s_{-1}])$

In this case, we are providing the LLM with a paired set of a previous transcript with its corresponding true scores. We compared the performance of this case to the similar models where we only supplied the previous transcript, the previous score, or neither:
- Model 0-shot: $\hat{s}_0 = f(x_0)$
- Model 0-shot+1-score: $\hat{s}_0 = f(x_0, s_{-1})$
- Model 0-shot+1-transcript: $\hat{s}_0 = f(x_0, x_{-1})$

As a benchmark control, we also compared the performance to simply substituting the previous score for the current one:
- Model last_score: $\hat{s}_0 = s_{-1}$

Finally, in the set of 45 patients with 3+ data points, we constructed an analogous set of model variants, such as:
- Model 2-shot: $\hat{s}_0 = f(x_0, [x_{-1}, s_{-1}], [x_{-2}, s_{-2}])$

## Results

*Overall Accuracy*

The mean total BPRS of patients within this sample was 38, close to other reported mean BPRS scores in outpatients with psychotic disorders (e.g. 35-43)[27,28]. The LLM provided sensible explanations and numerical predictions between 1-7 for each subscore for every case presented (Supplementary Materials). For the PSYCHS transcripts, the mean predicted total BPRS was 36, which was not significantly different from true mean ($p = 0.14$, $n = 374$, Mann-Whitney U test, Figure 1A), the Pearson correlation between predicted and true total scores was $r = 0.58$, the ICC was 0.73, the median concordance was 0.84, with 3 subscores with concordance less than 0.75.

We compared these to the inter- and intra-rater reliability measures reported by Hafkenscheid *et al.* (1993)[25], which primarily reported a median concordance of 0.83, also with 3 subscores with concordance less than 0.75 (Table 1). It also reported a longitudinal intra-rater Pearson correlation of 0.62 and an inter-rater ICC of 0.70 (of unknown specific ICC variant).

Open interview transcripts on the other hand had poorer accuracy, with an underpredicted mean total BPRS of 28, which was significantly different from true mean ($p < 0.001$, $n = 238$, Mann-Whitney U test, Figure 1B), the Pearson correlation between predicted and true total scores was $r = 0.39$, the ICC was 0.42, the median concordance was 0.88, with 5 subscores with concordance less than 0.75.

*Subscore Breakdown*

We examined the accuracy of the LLM's predictions of the 24 specific subscores that comprise the BPRS (Figure 2A). Using the same concordance threshold of 0.75 used by Hafkenscheid *et al.*, we found that the three subscores not accurately predicted from the PSYCHS interviews were the anxiety, depression, and conceptual disorganization subscores. Anxiety and depression were underestimated, whereas conceptual disorganization was overestimated. In the open interviews, which had a general underestimation bias, all five inaccurate subscores (anxiety, depression, guilt, suspiciousness, and hallucinations) were underestimated. Based on factor structures from a previous meta-analysis[29], these underestimated subscores tended to fall within the Affective and Positive Symptom factors (Figure 2B).

More generally, the BPRS classifies the first 14 subscores as self-reported subscores, which rely on the patient's subjective description[8]. Three of those 14, as well as the last 10 subscores are considered clinician-observed symptoms, which rely on the interviewer's assessment. Based on the Pearson correlation coefficients for the true vs. predicted individual subscores, we found that as a whole, the LLM was more accurate in predicting self-reported symptoms ($p=0.0001$, Mann-Whitney U test across symptoms, Figure 2C). This remained significant when subsetting to only the PSYCHS interviews ($p=0.0003$) but not the open interviews ($p=0.19$). Similarly, when comparing PSYCHS to open interviews, the self-reported scores were more accurately predicted from the PSYCHS transcripts ($p=0.002$, Mann-Whitney U test across symptoms, Figure 2D), whereas there was no difference for the observed scores ($p=0.92$).

*Foreign Language Interviews*
For our primary analysis, we only included English-language transcripts. However, given the ability of LLMs to work with other languages, we also investigated its behavior when using the same English-language prompt but supplied with a foreign language transcript (most of which were in either Spanish or Korean). The LLM performed comparably well on these foreign transcripts, with PSYCHS interviews demonstrating a Pearson correlation between predicted and true total scores of r = 0.54, the ICC was 0.70, the median concordance was 0.89, with 3 subscores with concordance less than 0.75 (Figure 3A). Open interviews also resulted in accurate scores, with a Pearson correlation of r = 0.53, the ICC was 0.63, the median concordance was 0.91, with 4 subscores with concordance less than 0.75.

The output explanations were generally written in English, but for some of the Spanish transcripts, the explanations were all written in Spanish (9 out of 57 cases). Other times, the explanation was in English, but would quote the patient in the foreign language as evidence to support the explanation (e.g. "The patient's mood was described as '보통' (normal) without specific worries or autonomic symptoms."), seamlessly integrating knowledge of foreign languages into its assessment (Figure 3B).

To investigate if transcripts from different countries and languages resulted in biased assessments, we compared the two foreign languages with the most examples (Spanish and Korean). Similar to the results with the English transcripts, both languages showed an underestimation of BPRS total score based on open interviews (Difference between true and predicted – Spanish: –4.8 ± 1.3, Korean: –6.9 ± 1.6, vs. English: –10.4 ± 0.6, SEM), but accurately estimated those based on PSYCHS interviews (Difference between true and predicted – Spanish: 0.8 ± 1.6, Korean: –2.3 ± 1.7, vs. English: –1.3 ± 0.4, SEM). The underestimated, low concordance subscores in these two languages were similar to the ones found in the English cohort (Spanish: anxiety, depression, blunted affect; Korean: anxiety, guilt).

*Longitudinal data*
Given the strengths of the model in zero-shot learning, we then utilized the longitudinal nature of this dataset to explore the performance of the LLM when given individualized examples of previous interview transcripts and scores. Our dataset had 153 patients with at least 2 timepoints with paired transcripts and scores, and 45 patients with 3 timepoints. We tested the LLM's ability to predict the most recent score based on the most recent transcript as well as data from previous timepoints. To understand the impact of including different components of past data, we tested providing the LLM with several different subsets of the past data (Figure 4A, see Methods for a detailed description of these models).

In the set of 153 patients with at least 2 timepoints, 1-shot learning (in which we provided both a previous score and transcript) resulted in the most accurate predictions, with an RMSE of 6.32 (Figure 4B). Both the 1-shot and 0-shot+1-score models performed better than the control case of using the last score, which had an RMSE of 7.19. The 0-shot and 0-shot+1-transcript models performed worse than the control case, highlighting the importance of supplying a previous score.

We then repeated this analysis in the 45 patients with 3 available timepoints. The top performing model was now the 2-shot learning model, in which two pairs of previous scores and transcripts were provided to the LLM (Figure 4C). Analogously to the 2-timepoint case, all models which incorporated at least one prior score and the current transcript were better performers than the control case of using the last score. Together, these findings demonstrate the trends that supplying additional previous scores or matched scores with transcripts helps model performance, whereas only supplying additional previous transcripts does not (Figure 4D-F).

**Discussion**

In this study, we demonstrate that LLMs can infer BPRS symptom severity scores from clinical interviews obtained in individuals at clinical high risk for schizophrenia. Despite being applied in a zero-shot configuration, with no model fine-tuning and no task-specific examples, the model's estimates from PSYCHS transcripts achieved concordance and ICC values that are similar to human inter- and intra-rater reliabilities. To our knowledge, this is the first application of an LLM for psychosis assessment and particularly for the CHR population. These findings extend prior work with LLMs in depression (i.e. with the DAIC-WOZ dataset) into other psychiatric conditions, showing that automated assessments may be feasible for various complex, nuanced metrics across the spectrum of mental health disorders.

Performance was typically better when the model was supplied the semi-structured PSYCHS interviews compared with the free-form open interviews. This discrepancy may be explained via two main factors. First, PSYCHS interviews systematically elicit information that overlaps with BPRS symptom domains such as hallucinations and unusual beliefs, whereas open interviews greatly vary in content type and interview length. Second, many BPRS items rely on self-reported descriptions of symptoms such as anxiety, guilt, and depression, which depend on explicit patient statements more likely to appear in a semi-structured context. When these reports were absent in open interviews, the model tended to underestimate the symptom scores, most notably for affective and positive symptoms which were relatively prevalent in this population. Notably, the PSYCHS interviews were not specifically designed around the BPRS, and were not shown to the model during training, suggesting that the LLM does not require exact overlap of symptom domains in order to infer symptom severity. Self-reported symptoms were typically more accurately predicted than observed symptoms, especially in PSYCHS interviews. We speculate that this is because the medium of interview transcripts preferentially carries information regarding questions the patient was asked about, rather than features that would require observations of the patient's movements or mannerisms. Future multimodal approaches that utilize video or direct audio data may have improved detection of these observed features.

Compared to a human rater approach, using LLMs has some potential advantages. The model proficiently translated, contextualized, and assessed non-English transcripts, demonstrating a latent cross-language capability that could standardize symptom assessment across multi-national consortia like AMP-SCZ and reduce culturally or linguistically driven rater variance. It also could naturally incorporate longitudinal context (earlier interviews and their ground-truth scores) through n-shot prompting, which substantially improved performance over the zero-shot approach. This illustrates how LLMs can be adapted to an individual's specific treatment course, a longstanding goal in precision psychiatry.

There are several limitations and open challenges to our work.

*Prompt sensitivity*
Preliminary testing revealed that seemingly innocuous wording changes could strongly impact scoring, a known phenomenon with LLMs[30]. More systematic prompt-engineering frameworks will therefore be essential before clinical deployment. Of note, it was found that supplying the BPRS interviewer manual as part of the prompt was critical for the functioning of the LLM. Without this, the LLM could not even reliably list the 24 items of the BPRS. Other work on LLMs in the depression space typically do not need to supply the definitions for the PHQ-8, likely as that scale is better represented in LLMs' training data.

*Choice and availability of clinical scales*
Ideally, the assessment of CHR is best performed through specific instruments designed to measure prodromal psychosis, including the SIPS, CAARMS, and PSYCHS scores. Although AMP-SCZ collects these other CHR instruments, the data available from the current release is too incomplete to permit the same evaluation. Furthermore, the PSYCHS instrument is too new to have substantial outside data, and in fact, we found LLM is not even familiar enough with the PSYCHS to be able to list out its component items. Consequently, we cannot yet determine whether LLMs can assess these disease-specific scales, which may be more sensitive for assessing the conversion to psychosis than the BPRS. However, the BPRS and its components have already shown utility in predictive modeling in the CHR population[9–14], which is encouraging for its future integration into outcomes-based risk calculators. Promisingly, certain key symptoms noted by studies such as NAPLS that are associated with conversion to psychosis (e.g. suspiciousness[3], hallucinations[31]) were predicted with high accuracy by the LLM.

*Sparse longitudinal coverage*
Only 45 participants had three usable time-points, and none had more than three. Larger longitudinal sequences would be necessary to probe whether performance improvements saturate or continue to accrue with additional context, and to examine the model's capacity to detect within-subject changes rather than absolute severity alone.

*Ground-truth reliability constraints*
Human BPRS inter- and intra-rater reliability data are limited to a single study. Additionally, this study did not specify certain methodological details such as which ICC variant it computed. Because the performance of the LLM in predicting gold-standard clinician ratings cannot exceed this human reliability constraint, future work should involve concurrent, multiple-rater re-scoring of a subset of transcripts to establish an updated benchmark.

As the AMP-SCZ project continues to release additional data, the present methodology could be reapplied to derive SIPS, CAARMS, PSYCHS or newly developed digital phenotyping metrics. The ease with which LLMs can integrate multiple modalities also opens the possibility of including other data forms from the project (e.g. passive smartphone sensor data, EEG, or neuroimaging reports) alongside interview transcripts, potentially yielding richer and more prognostically useful summary scores. Fine tuning, especially with larger and more complete data sets, also presents a promising avenue for continued model performance improvements. The eventual goal of these efforts would be to develop better risk stratification tools predicting the conversion of CHR patients, allowing for improved management and earlier intervention for these individuals. On the implementation side, integrating real-time LLM scoring such as this approach into psychiatry platforms could also give clinicians immediate, structured data without extending visit lengths.

**Conclusion**

Our findings are a proof of concept that contemporary LLMs can approximate expert human ratings on a complex, multi-item psychosis scale using nothing more than routine clinical text, even across different languages. By lowering the logistical barrier to structured symptom assessment, this approach could accelerate research, harmonize multi-site collaborations, and ultimately support earlier and more precise intervention for individuals at risk of schizophrenia.

**Tables and Figures**

**Table 1: Performance of LLM zero-shot approach in predicting BPRS scores.** For comparison, the results from a human intra- and inter-rater reliability study (Hafkenscheid *et al.* 1993) are also shown. On multiple different metrics of accuracy, the LLM assessments from PSYCHS transcripts (both English and foreign language) approach human inter- and intra-rater reliability.

|  | Pearson *r* | Median Concordance | Concordance #subscores<0.75 | ICC |
|---|---|---|---|---|
| Hafkenscheid *et al.* 1993 | 0.62 | 0.83 | 3 | 0.70 |
| LLM-English-psychs | 0.58 | 0.84 | 3 | 0.73 |
| LLM-English-open | 0.39 | 0.88 | 5 | 0.42 |
| LLM-Foreign-psychs | 0.54 | 0.88 | 3 | 0.70 |
| LLM-Foreign-open | 0.53 | 0.91 | 4 | 0.63 |

**Figure 1: A zero-shot approach in predicting BPRS scores from interview transcripts with LLMs.** LLMs can accurately predict total BPRS scores from (A) semi-structured PSYCHS transcripts, though perform worse and underestimate total scores on (B) unstructured open transcripts.

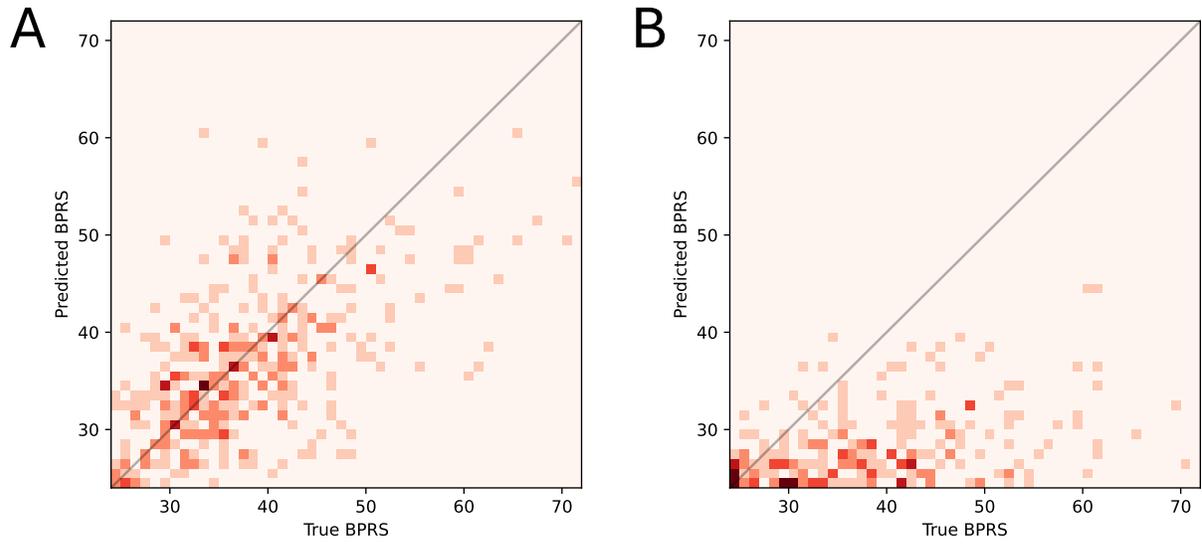

**Figure 2: BPRS subscores derived from LLM predictions.** (A) Subscore means from open and PSYCHS transcripts compared to their true means, as well as their Pearson correlation coefficients. (B) Grouping of subscores totals into core factor categories. (C) LLM performance as measured by Pearson correlation was higher for self-reported features compared to observed features. (D) This effect appears to stem mainly from the higher accuracy of PSYCHS interviews to assess self-reported features.

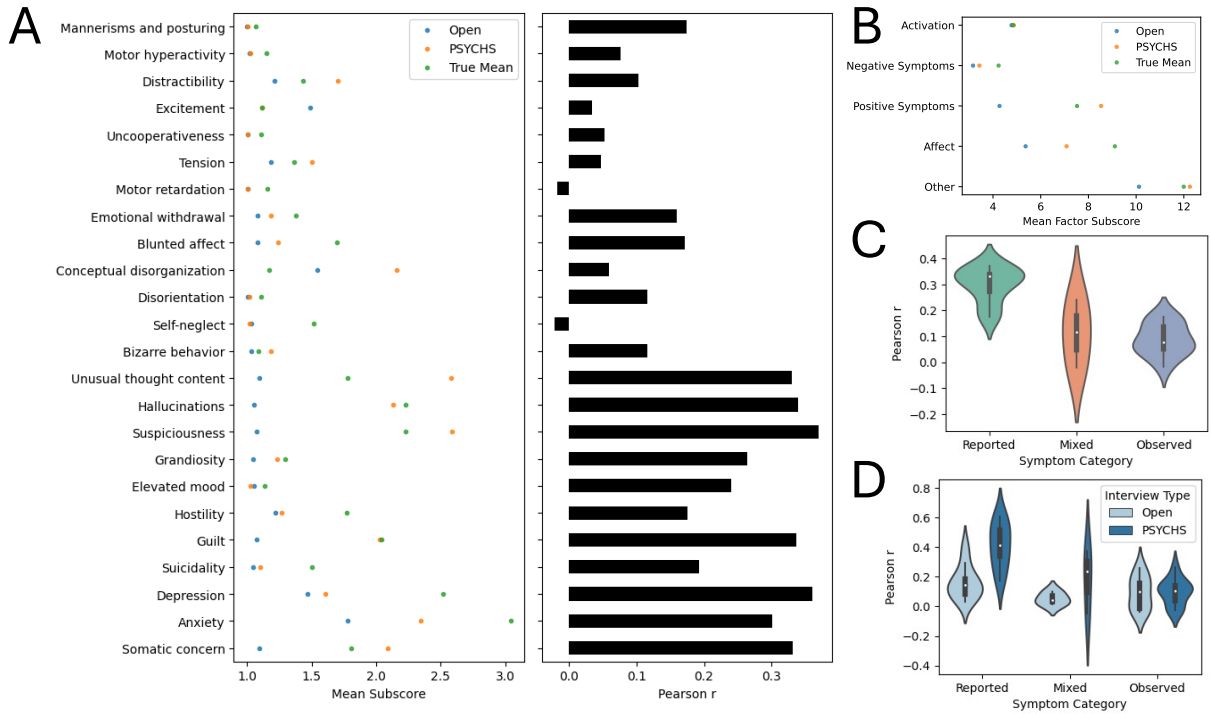

**Figure 3: Interview transcripts in foreign languages.** (A) LLMs can also predict total BPRS scores accurately from foreign language transcripts in a zero-shot approach. (B) In the explanations of its score predictions, LLMs seamlessly integrate foreign words as evidence.

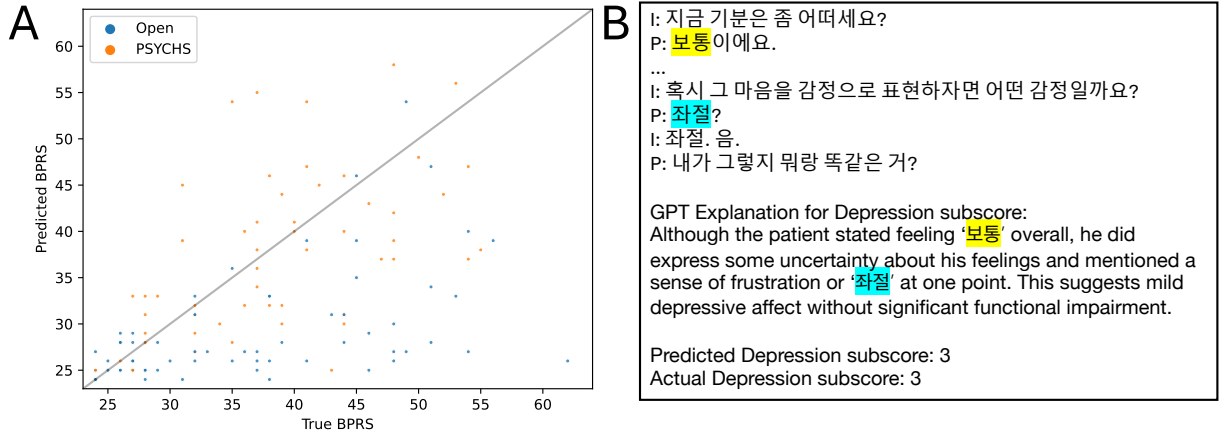

**Figure 4: Performance improvements with longitudinal data.** (A) Schematic of different models which incorporate previous data points from the same patient into the LLM's context window. These models in which the LLM was supplied varying combinations of previous transcripts and previous scores were compared to the zero-shot model used for the previous figures, and a longitudinal control case, in which the previous score was directly carried forward as the prediction for today's score. Comparison of performance across all models for the data subset with two or more time points (B) and for the data subset with three time points (C). (D) Within the cohort of 45 patients with three time points available, the 1-shot and 2-shot approaches both had significantly higher performance compared to the zero-shot approach. (E) Adding previous transcripts to the zero-shot model did not improve performance, but supplying previous scores did increase accuracy (F). Colors denote specific highlighted models - blue: the pure n-shot models, green: longitudinal controls, grey: all others.

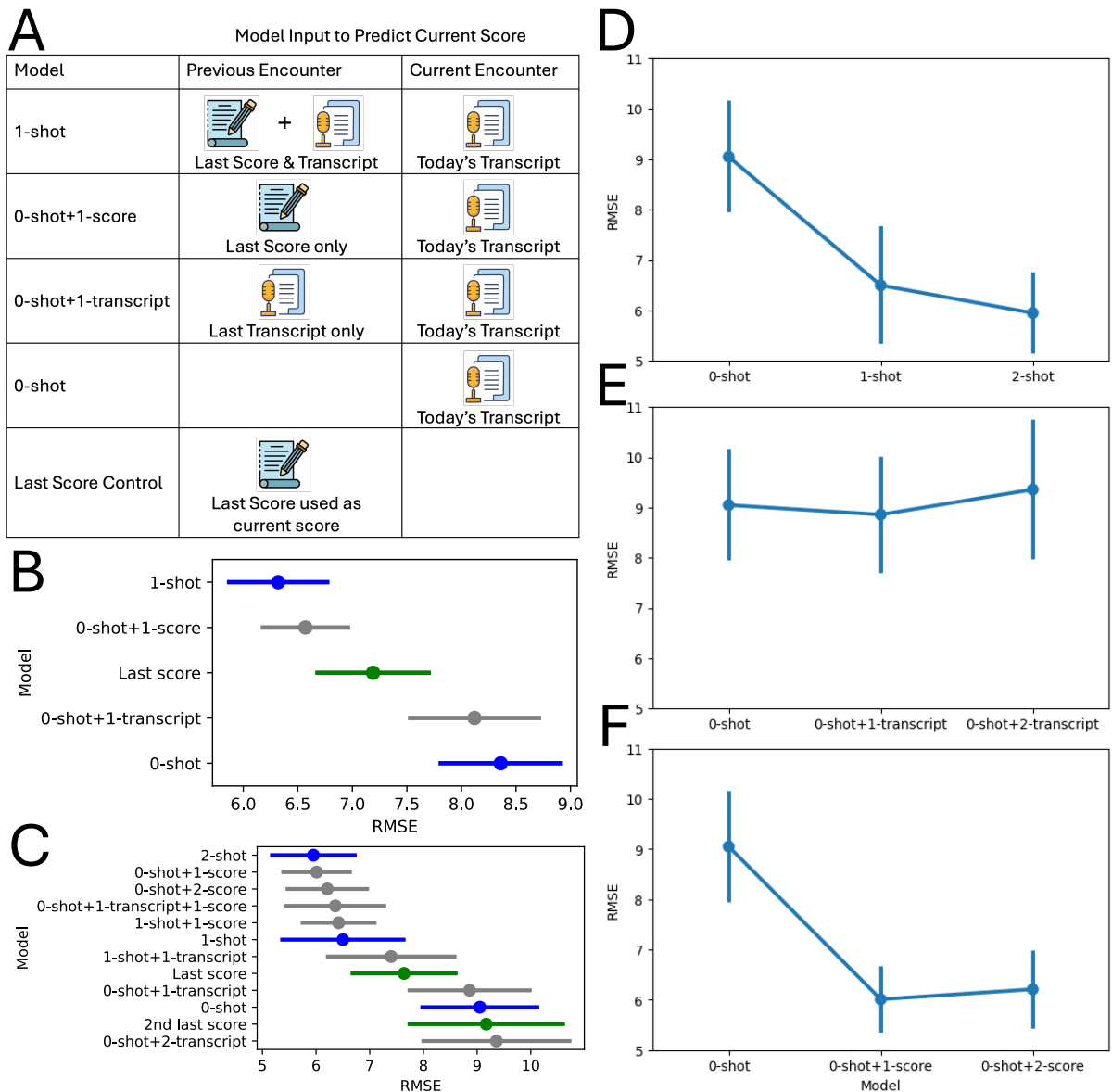